\documentclass[runningheads]{llncs}
\usepackage[T1]{fontenc}
\usepackage{graphicx}
\usepackage{multirow}
\usepackage{amssymb}
\usepackage{graphicx}
\usepackage{wrapfig}
\usepackage{booktabs}
\usepackage{amsmath}
\usepackage{soul}
\usepackage{pifont}
\usepackage{amssymb}
\usepackage{bm}
\usepackage{color}
\usepackage[dvipsnames]{xcolor}
\usepackage{gensymb}
\usepackage{orcidlink}
\usepackage{amsmath}
\usepackage{lineno}
\usepackage{comment}
\usepackage{subcaption}
\usepackage{float}
\usepackage{placeins}

\usepackage{placeins}

\begin{document}
%\linenumbers

\title{Global–Local Feature Decoding with Adapter-Guided SAMv2 for Salient Object Detection}

% TODO REVIEW: If the paper title is too long for the running head, you can set
% an abbreviated paper title here. If not, comment out.
\titlerunning{Global-Local Decoding with SAM2 for SOD}

\author{
  Morteza Moradi\inst{1} \and
  Mohammad Moradi\inst{1} \and
  Simone Palazzo\inst{1} \and
  Ali Borji\inst{2}    \and
  Concetto Spampinato\inst{1}
}
\authorrunning{Moradi et al.}

\institute{University of Catania, Catania, Italy \\
\email{mohammad.moradi@phd.unict.it\\\{morteza.moradi, simone.palazzo, concetto.spampinato\}@unict.it} \and
Quintic AI, San Francisco, CA, USA\\
\email{aliborji@gmail.com}}

\maketitle              
\begin{abstract}
Salient Object Detection (SOD) remains an essential yet underexplored task in the era of large-scale vision models. Although foundation models like SAM exhibit strong generalization, their potential for SOD is not fully realized, and training or fully fine-tuning them is computationally expensive and prone to overfitting under limited data. To overcome these challenges, we introduce GLASSNet, a Global–Local feature decoding framework that uses SAMv2 as a frozen encoder paired with a lightweight, spatially aware convolutional adapter—reducing learnable encoder parameters by over 97\%. To enhance saliency quality, GLASSNet employs a dual-decoder architecture: one decoder captures global, long-range semantics with an expanded receptive field, while the other captures fine local details such as edges and textures. Fusing these complementary cues yields saliency maps that combine global coherence with local precision, producing accurate final masks. Extensive experiments on standard SOD and camouflaged object detection benchmarks show that GLASSNet surpasses state-of-the-art methods, demonstrating the power of frozen foundation models combined with targeted adaptation and global–local decoding.
\end{abstract}

\keywords{Salient Object Detection \and Segment Anything Model \and Dual-stream Decoder \and Adapter Network \and Camouflaged Object Detection}

\section{Introduction}
\label{sec:intro}
Salient Object Detection (SOD) \cite{borji2019salient}—the task of segmenting visually dominant objects—has progressed from handcrafted features \cite{liu2010learning} to Transformer-based models \cite{wang2022tf}, greatly improving accuracy and robustness. These advances have expanded SOD applications in various areas such as medical imaging and autonomous navigation.

\begin{figure}[t]
    \centering
    \includegraphics[width=0.7\linewidth]{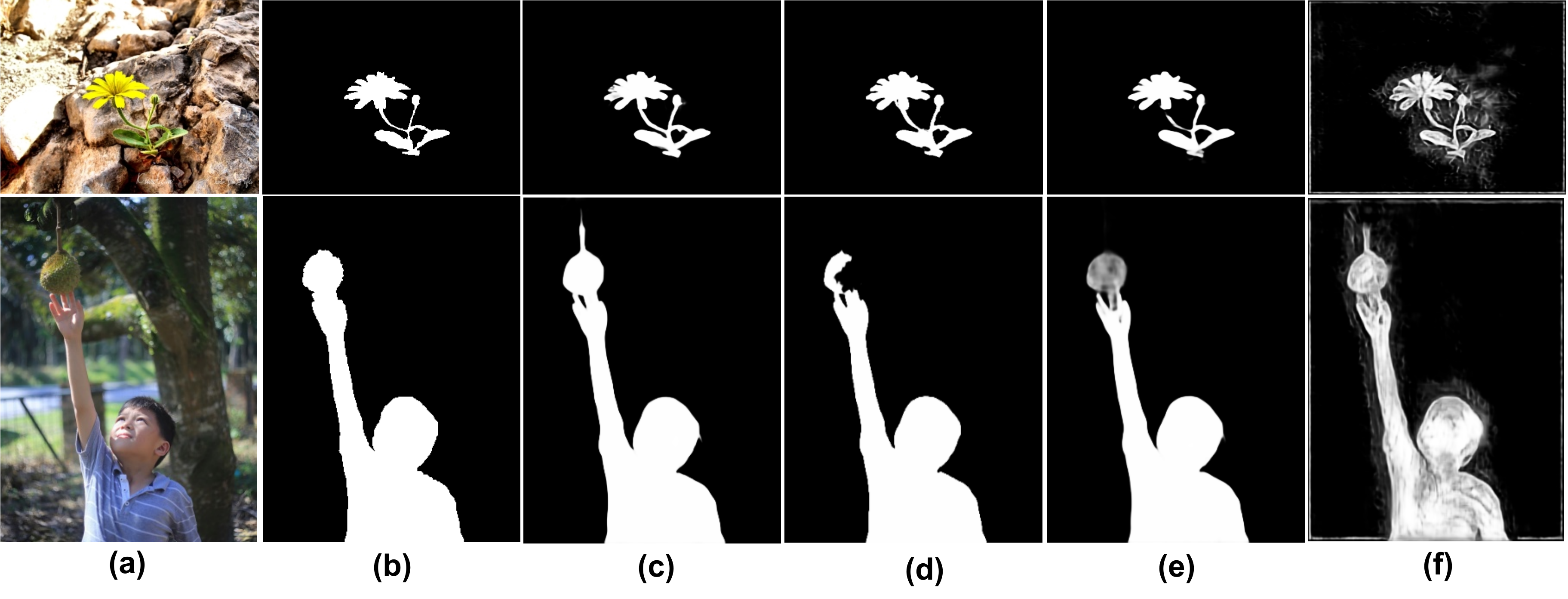} % replace with your actual file path
    \caption{Intermediate outputs of GLASSNet. Columns (a)–(f) show the input images, ground truth, model output, and outputs of the local, global, and mid-level decoders, illustrating each component's contribution to the final saliency map.}
    \label{fig:component_outputs}
\end{figure}
Despite this progress, many SOD methods still struggle to generalize, showing degraded performance in diverse or domain-specific scenarios.

Recent large-scale vision foundation models have reshaped saliency modeling \cite{zhou2024vision,moradi2024salfom} and broader visual understanding \cite{kazmierczak2025explainability}, offering far stronger generalization and addressing key limitations of CNNs and early transformers. These capabilities enable more reliable and adaptable SOD systems.

Among recent image foundation models, the Segment Anything Model (SAM) \cite{kirillov2023segment} stands out as a strong backbone for salient object detection. Its object-centric pretraining, ability to generate sharp segmentation masks, and robust generalization across domains \cite{kirillov2023segment} align well with the core objectives of SOD. Unlike conventional models—which often require extensive task-specific annotations or struggle to generalize—SAM provides zero-shot, instance-level masks, streamlining the initial mask generation process in saliency workflows. Building on this capability, several recent studies have explored the use of SAM for SOD tasks \cite{cui2023adaptive,liu2024weakly}.

However, SAM is not a plug-and-play solution. As a general-purpose model, it requires careful adaptation to meet the specific demands of SOD, and performance depends on how well downstream components interpret and refine its outputs. While SAM lowers the entry barrier, its effectiveness in saliency detection still hinges on proper integration and architectural tuning.

Specifically, SAM can produce fragmented or imprecise masks in scenes with subtle boundaries \cite{ji2024segment}, low contrast \cite{ji2023sam}, fine and intricate structures \cite{zhang2024quantifying} (e.g., branches or veins), or transparent and reflective surfaces \cite{han2023segment}. Relying solely on its raw outputs can propagate these limitations, making SAM a strong starting point but not a complete solution for SOD.

To address these issues, and based on the hypothesis that with more calibrated modulation and targeted network design SAM can be even more effective for salient object detection, we propose \textbf{G}lobal–\textbf{L}ocal feature decoding with \textbf{A}dapter-guided \textbf{S}AM v2 for \textbf{S}alient object detection (GLASSNet), a novel SOD model that employs SAMv2 \cite{ravi2024sam} as a frozen encoder backbone. In our framework, the SAMv2 encoder generates high-resolution, object-aware feature maps that serve as a rich initial representation of the input image. Instead of fine-tuning the large and computationally expensive SAM backbone, we keep it entirely frozen and introduce a lightweight, spatially-aware convolutional adapter to modulate and adapt its features for the task. This adapter is designed to preserve spatial continuity and object boundaries while aligning the SAM-derived features with the saliency task. Its convolutional nature allows it to better capture local texture and edge information compared to MLP-based alternatives, while keeping the model efficient.

%\textcolor{red}{Following this, the adapted feature maps are passed to a dual-decoder architecture, with each branch specializing in a complementary aspect of saliency detection. The global decoder captures semantic dependencies and high-level object interactions over a wide spatial field, using context-aggregation attention, dilated convolutions, and multi-scale context reasoning to estimate object importance in the broader scene. In parallel, the local decoder emphasizes fine-grained cues—boundaries, textures, and small structures—through locality-aware attention, consecutive convolutions, and shallow feature fusion. Their outputs are finally merged by a feature aggregation module that integrates global and local cues into a unified saliency map.}
The adapted feature maps are processed by a dual-decoder architecture. The global decoder captures high-level semantic dependencies using context-aggregation attention, dilated convolutions, and multi-scale reasoning, while the local decoder focuses on fine-grained cues—boundaries, textures, and small structures—via locality-aware attention and shallow feature fusion. A feature aggregation module then combines their outputs into a unified saliency map.

Fig~\ref{fig:component_outputs} illustrates the outputs of each component within our model, highlighting their individual contributions to the final saliency detection.

To summarize, our key contributions are:

\begin{itemize}
    \item We guide the frozen SAMv2 encoder using a lightweight, spatially-aware adapter network to adapt its general-purpose representations to the SOD task effectively.
    
    \item We design a dual-stream decoder that integrates global semantic context and local structural cues to enhance saliency detection.
    
    \item Our model not only surpasses state-of-the-art SOD methods but also outperforms recent task-specific models in camouflaged object detection, demonstrating strong generalization across domains.
\end{itemize}

\begin{comment}
The remainder of the paper is structured as follows. Section 2 reviews related work and summarizes recent advances. Section 3 presents the architectural design of the proposed model. Section 4 reports and analyzes experimental comparisons between our model, GLASSNet, and state-of-the-art methods on salient and camouflaged object detection benchmarks.
\end{comment}

\vspace{-10pt}
\section{Related Works}
\label{sec:Relatedworks}

From a broad perspective, our proposed SOD model builds upon two key design choices: leveraging a vision foundation model as the encoder backbone, and employing a two-stream local-global decoder architecture. In this section, we briefly review prior works that adopt similar strategies and highlight the conceptual and architectural distinctions of our approach.

\textbf{Vision Foundation Model-Driven Saliency Modeling} 
Recent advances in vision foundation models \cite{wang2022omnivl}, pretrained on massive datasets, have greatly improved model generalizability in visual understanding. These models have advanced saliency modeling for images \cite{zhang2024category} and videos \cite{moradi2024salfom,zhou2024vision}. In this context, foundation models—particularly SAM, with its object-centric and instance-level masks—have become promising for SOD, though existing SAM-based methods mainly depend on prompt-based usage or task-specific adapters.

Early work such as SSOM \cite{cui2023adaptive} fine-tunes SAM using AdaLoRA while keeping parameter updates lightweight. Liu and Huang \cite{liu2024weakly} integrate SAM into a two-stage weakly supervised SOD pipeline using box prompts, where SAM-generated masks create pseudo-labels that are progressively refined. SSFam \cite{liu2025ssfam} adapts SAM for scribble-supervised SOD across multiple modalities (RGB, RGB-D, RGB-T, V-D-T), introducing modality-aware modulators and using scribbles as prompts during training, with a siamese decoder enabling prompt-free inference.

\textbf{Incorporating local and global contexts}
Integrating local and global information has long been a central strategy in SOD, though these concepts are modeled differently across works. In \cite{tong2015salient}, global cues come from bottom-up saliency maps, while local cues are derived from top-down maps. The context-aware attention network in \cite{ren2020salient} captures semantic relationships between each pixel and both its local and global surroundings. Other methods learn local-to-global features to exploit fine-grained structure for global saliency inference \cite{feng2022local,ji2022lgcnet}. While many models fuse local and global cues within a single architecture, some explicitly adopt two-stream designs to separately model each type of context \cite{yun2022selfreformer}. 

In this work, we use a dual-stream decoding network, with each branch responsible for a different type of contextual information. One decoding branch captures global, long-range semantic and conceptual dependencies inherent in the image (and among pixels of salient objects) by expanding the receptive field, while the other focuses on fine-grained local structural features—including edges, textures, and boundaries—to distinguish the object(s) from the surroundings. 

\section{Methodology}
\label{sec:methodology}
\subsection{Overview}
\label{sec:overview}
\begin{figure}[t]
    \centering
    \includegraphics[width=0.7\linewidth]{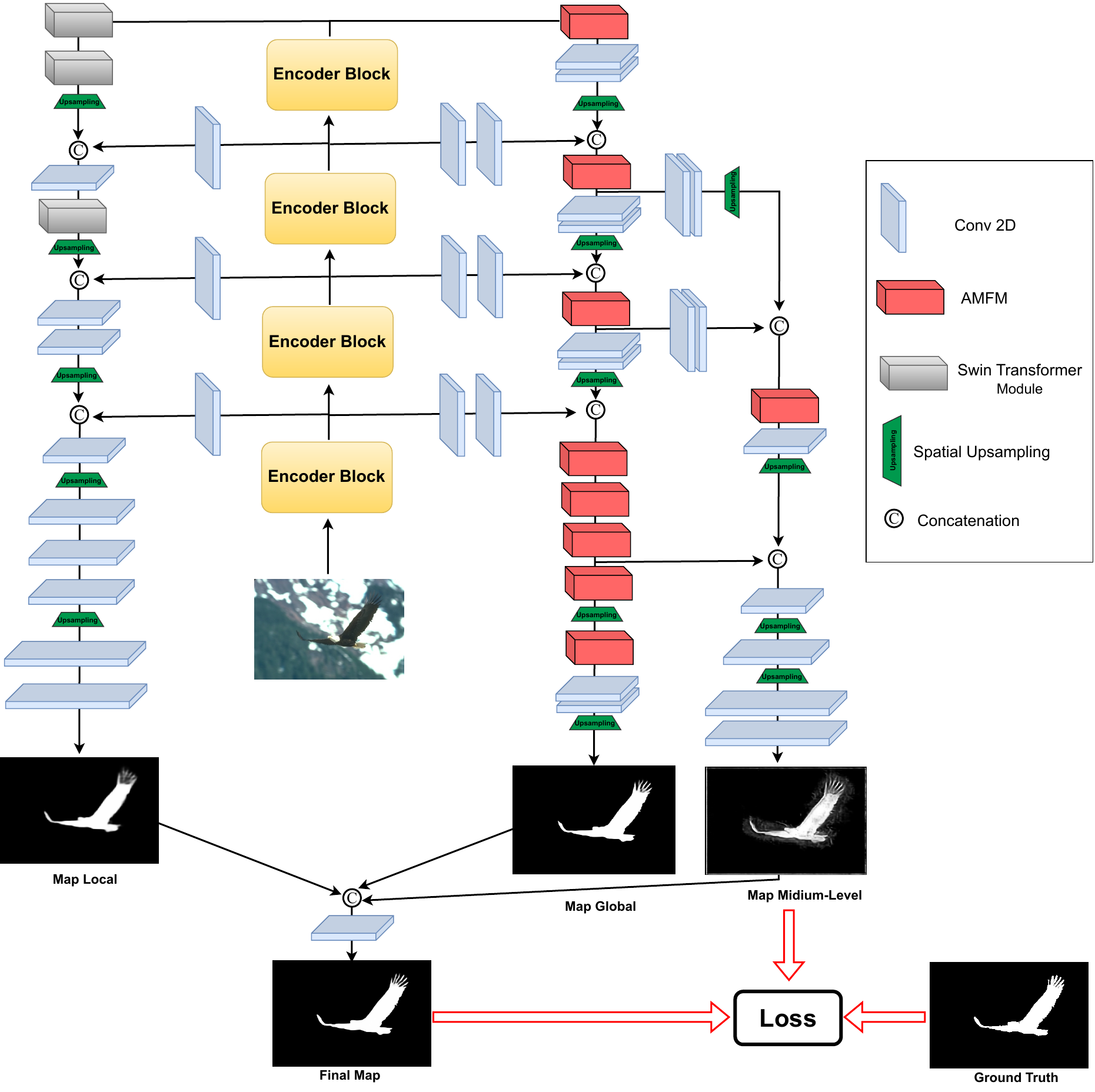}
    \caption{Schematic overview of GLASSNet for saliency detection. A frozen SAMv2 encoder is adapted with a lightweight convolutional adapter and a multi-branch decoder: local for fine details, global for contextual features, and medium-scale for balance. Their fused outputs produce accurate multi-scale saliency maps.}
    \label{fig:arch}
\end{figure}

\vspace{-5pt}
The architecture of the proposed SOD model, \textbf{GLASSNet}, is shown in Fig.~\ref{fig:arch}. We adopt the image encoder of \textbf{SAMv2}—excluding its prompt encoder and mask decoder—as a frozen backbone to reduce computation and avoid overfitting on limited SOD data.

To adapt SAMv2’s features to the saliency task, we introduce lightweight convolutional adapters at the input of each encoding layer. These adapters modulate features in a task-specific manner without modifying the original backbone. Unlike common MLP-based adapters, our convolution-based design better preserves spatial detail, which is crucial for dense prediction.

The encoder–decoder structure follows a U-shaped architecture, with decoder modules progressively fusing multi-scale features from the encoder. To capture different contextual cues, we design two decoders: one for fine-grained local details and another for long-range global dependencies. Additionally, a sub-branch derived from the global decoder captures mid-level context. The outputs from all three branches are fused to generate the final saliency map.

\subsection{Encoder Structure}
\label{sec:encoder}

In our network, we employ only the image encoder of \textbf{SAMv2}~\cite{ravi2024sam} as the feature extractor, omitting its prompt encoders and mask decoder. This decision is driven by the strong generalization capability of SAMv2’s vision backbone, which performs effectively across diverse vision tasks.

At the core of the image encoder lies \textbf{HiERA} (Hierarchical Encoder with Relative Attention)~\cite{ryali2023hiera}, a transformer-based architecture designed for efficient, scalable segmentation. Unlike the single-scale ViT encoder in SAM v1, HiERA introduces a hierarchical design that extracts multi-scale features, making it naturally compatible with encoder–decoder frameworks like U-Net.

We adopt the pretrained HiERA-L variant, which outputs four levels of hierarchical feature maps given an input image $\mathbb{R}^{3 \times H \times W}$. These features are denoted as $X_i \in \mathbb{R}^{C_i \times \frac{H}{2^{i+1}} \times \frac{W}{2^{i+1}}}$, where $i \in \{1, 2, 3, 4\}$ and $C_i \in \{144, 288, 576, 1152\}$.

\begin{figure}[t]
    \centering
    \includegraphics[width=0.6\linewidth]{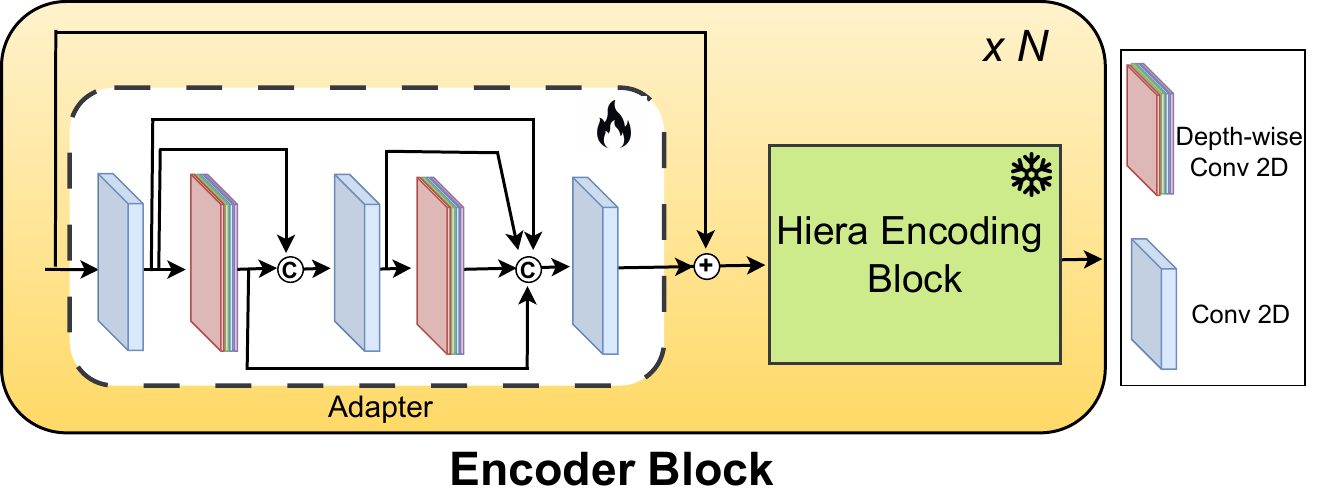} 
    \caption{A lightweight adapter is embedded into the SAMv2 image encoder. It uses point-wise and depthwise convolutions to capture multi-scale local features, fuses them via concatenation and residual connections, and enhances representations with minimal added parameters for efficient fine-tuning.}
    \label{fig:adapter}
\end{figure}

To adapt the frozen SAMv2 encoder to the saliency detection task, we introduce a lightweight convolutional adapter at the input of each encoding stage. Unlike typical MLP-based adapters, our design employs five densely connected convolutional layers (kernel size 3) combining standard and depthwise convolutions. This structure enhances spatial sensitivity and promotes feature reuse and gradient flow—critical for adapting frozen representations to dense prediction tasks. The adapter architecture is illustrated in Fig.~\ref{fig:adapter}.

\subsection{Decoding Structure and Workflow}
\label{sec:decoder}
\vspace{-5pt}

We design a dual decoder structure comprising to capture features across local, global, and intermediate (medium-scale) contexts.

\textbf{Local Decoder.}  
The local decoder focuses on fine-grained spatial details essential for precise saliency detection, such as boundary sharpness and foreground–background separation. It processes features from the deepest encoder layer alongside intermediate encoder outputs using a combination of Swin Transformer \cite{liu2021swin} blocks (depth 2, window size 1) and 2D convolutions. Swin Transformer blocks are used early for local refinement and channel recalibration, while later stages rely on 2D convolutions to maintain computational efficiency at higher resolutions. The decoder follows a progressive fusion scheme: features are upsampled and merged with shallower encoder outputs at each stage, with spatial resolution gradually doubled and channel dimensions reduced. This structure enables the network to reconstruct a high-resolution saliency map with strong local accuracy.

\textbf{Global Decoder.}  
To capture long-range dependencies and high-level semantics, we introduce a global decoder that progressively enlarges the receptive field. Its core component is the \textbf{Attentional Multi-Scale Feature Fusion Module (AMFM)} (Fig.~\ref{fig:module}), which begins with a Criss-Cross Attention \cite{huang2019ccnet} mechanism to model global relationships across horizontal and vertical dimensions. This is followed by three parallel convolutional branches: one standard and two dilated (with rates of 2 and 3), enabling multi-scale context aggregation. Outputs are fused via pixel-wise summation. Additionally, a depthwise separable convolution applied to a residual connection enhances channel-wise feature weighting. The feature decoding procedure in this branch is gradually started from deepest to shallowest encoder layers, integrating intermediate features while maintaining spatial abstraction. As illustrated in the network's architecture, the consecutive utilization of AMFM progressively enlarges the receptive field, allowing the model to capture broader contextual information. Additionally, upsampling layers at certain stages help restore the spatial resolution of the final saliency map. The final stages of this branch use convolutional layers to refine and generate the saliency map. 

%\begin{figure}[t]
    %\centering
    %\includegraphics[width=0.35\linewidth]{module.pdf} 
    %\caption{The AMFM captures global context using Criss-Cross Attention and fuses multi-scale features via parallel convolutions. Outputs are %summed pixel-wise, with a depthwise separable residual convolution refining channel features.}
%    \label{fig:module}
%\end{figure}
\noindent
\begin{minipage}{\linewidth}
    \centering
    \includegraphics[width=0.35\linewidth]{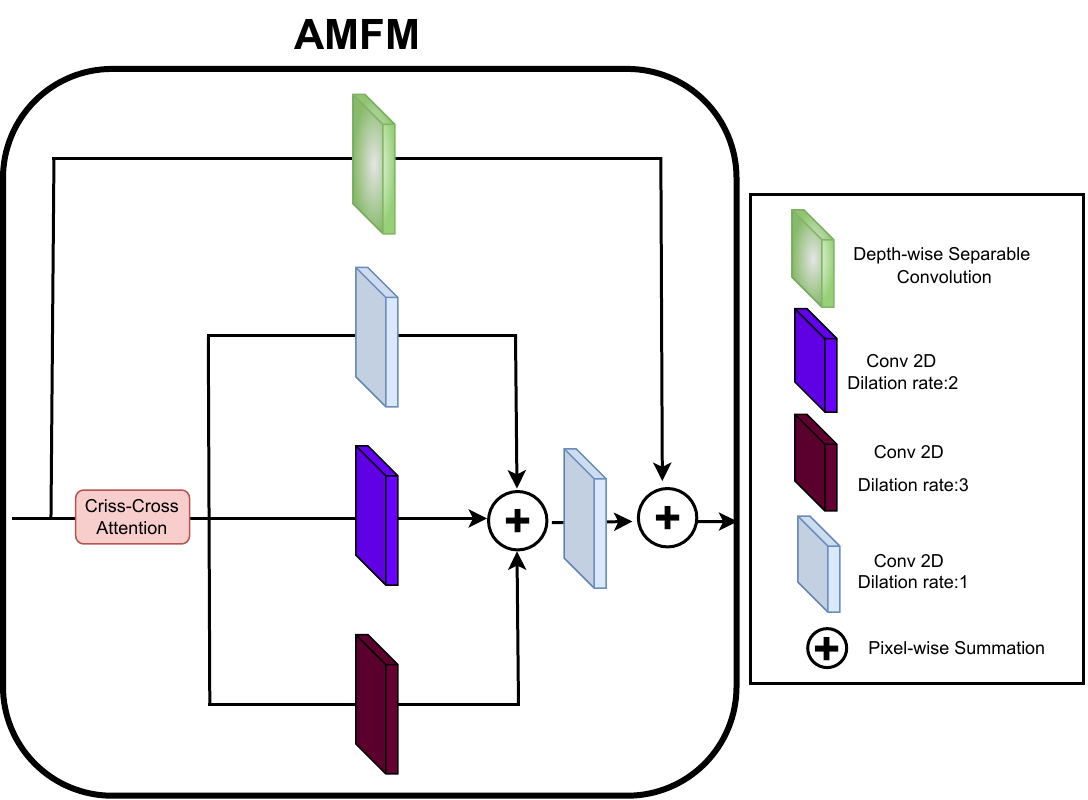}
    \captionof{figure}{The AMFM captures global context using Criss-Cross Attention and fuses multi-scale features via parallel convolutions. Outputs are summed pixel-wise, with a depthwise separable residual convolution refining channel features.}
    \label{fig:module}
\end{minipage}
\vspace{3pt}

\textbf{Medium-Scale Decoder.}  
To bridge the gap between local and global features, we add a lightweight sub-branch derived from the global decoder. This \textit{medium-scale decoder} uses a single AMFM modul and applies sharper channel reduction for efficiency. It also employs intermediate outputs from the global decoder and focuses on medium-range contextual reasoning. Though shallow, it benefits from globally refined features and complements both decoders by enhancing mid-level feature representation.

Finally, each decoder produces an individual saliency map, and these are fused to form the final output. In the decoder branches, all Conv2D layers use a kernel size of 3. A ReLU activation function and a Batch Normalization layer are applied after each convolutional layer, except for the final one. The parallel multi-scale decoding design enables the model to specialize in different contextual scopes, leading to more stable learning and accurate saliency detection across spatial scales.

\subsection{Training Objectives}
\label{sec:loss}
\vspace{-5pt}

%Given an input image $I \in \mathbb{R}^{H \times W \times 3}$ and its corresponding ground-truth saliency map $G \in \mathbb{R}^{H \times W}$, the objective of the model during training is to generate an estimated saliency map $S \in \mathbb{R}^{H \times W}$. The loss function is inspired by prior work~\cite{wei2020f3net} and is defined as a combination of two components, $\mathcal{L} = \mathcal{L}_{\text{wIoU}}(S, G) + \mathcal{L}_{\text{wBCE}}(S, G)$, where $\mathcal{L}_{\text{wIoU}}$ denotes the weighted Intersection over Union (IoU) loss and $\mathcal{L}_{\text{wBCE}}$ represents the weighted binary cross-entropy (BCE) loss. These terms are designed to guide the model toward generating saliency maps that align closely with the ground-truth annotations.

%To prevent the model from overfitting to either local or global features—typically captured by the local and global decoders—we exclude these outputs from the loss computation. Instead, we incorporate an auxiliary supervision term $\mathcal{L}_{\text{mid}}$, computed from the output of a complementary decoding branch associated with the global decoder. This mid-level supervision encourages the learning of representations that balance local detail and global context.

%The final training objective is therefore defined as $\mathcal{L}_{\text{total}} = \mathcal{L}(S_{\text{final}}, G) + \mathcal{L}(S_{\text{mid}}, G)$, where $S_{\text{final}}$ denotes the final saliency map and $S_{\text{mid}}$ denotes the saliency map generated by the complementary branch of the global decoder.

Given an input image $I \in \mathbb{R}^{H \times W \times 3}$ and its ground-truth saliency map $G \in \mathbb{R}^{H \times W}$, the model aims to predict an estimated saliency map $S \in \mathbb{R}^{H \times W}$. Following~\cite{wei2020f3net}, the loss is defined as 
$\mathcal{L} = \mathcal{L}_{\text{wIoU}}(S, G) + \mathcal{L}_{\text{wBCE}}(S, G)$, 
where $\mathcal{L}_{\text{wIoU}}$ is the weighted IoU loss and $\mathcal{L}_{\text{wBCE}}$ is the weighted binary cross-entropy loss, guiding the model to align predictions with the ground truth.

To avoid overfitting to local or global features, outputs from the local and global decoders are excluded from the loss. Instead, an auxiliary supervision term $\mathcal{L}_{\text{mid}}$, computed from a complementary branch of the global decoder, encourages learning representations that balance local detail and global context.

The final objective is $\mathcal{L}_{\text{total}} = \mathcal{L}(S_{\text{final}}, G) + \mathcal{L}(S_{\text{mid}}, G)$,
where $S_{\text{final}}$ is the final saliency map and $S_{\text{mid}}$ is the output of the complementary global branch.

\section{Experiments}
\label{sec:experiments}
\begin{figure*}[t]
    \centering
    \includegraphics[width=0.75\linewidth]{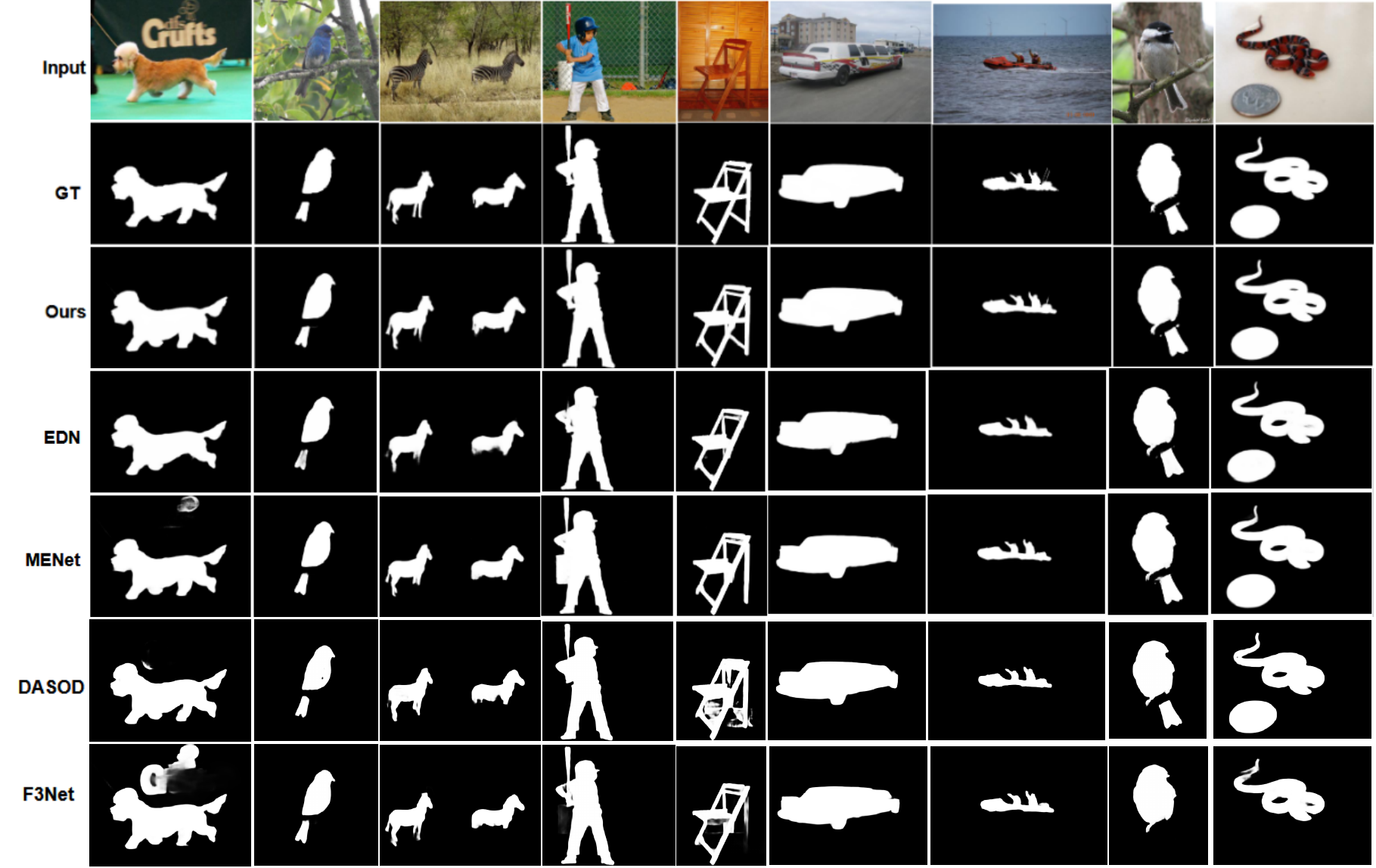}
    \caption{Qualitative comparisons show our model effectively reconstructs local and global features, producing saliency maps closely matching the ground truth.}
    \label{fig:map}
\end{figure*}

\subsection{Experimental Setup}

To ensure fair and consistent evaluation, we train our model on the training set of the DUTS-TE dataset~\cite{wang2017learning}, which contains 10,533 images. The trained model is then evaluated on five widely used salient object detection (SOD) benchmark datasets: the test set of DUTS-TE~\cite{wang2017learning} (5,019 images), HKU-IS~\cite{li2015visual} (4,447 images), DUT-OMRON~\cite{yang2013saliency} (5,168 images), ECSSD~\cite{yan2013hierarchical} (1,000 images), and PASCAL-S~\cite{li2014secrets} (850 images). Training was performed on an NVIDIA H100 GPU. The image encoder weights were initialized from the SAMv2-L model, while the remaining parameters of our network were randomly initialized. Input images were resized to \(352 \times 352\) pixels, and a batch size of 16 was used. The model was optimized using AdamW with a weight decay of \(1 \times 10^{-4}\) and a learning rate of \(5 \times 10^{-4}\). During training, the SAMv2 encoder parameters were frozen, and only the appended adapter layers were updated. The model was trained for up to 60 epochs. %Additionally, the model has a size of 1.311 GB and the average inference time per image is 0.048 seconds.

\subsection{Evaluation Metrics}
Following prior works \cite{ma2023boosting,asheghi2024dasod,khan2025pyramidal}, we evaluate SOD performance using four standard metrics: Mean Absolute Error (MAE), structural similarity ($S_m$), mean Enhanced-alignment Measure ($E_{mean}$), and maximum F-measure ($F_{\max}^{\beta}$).

MAE measures pixel-wise differences between a predicted saliency map $S$ and ground truth $GT$, computed as $ \text{MAE} = \frac{1}{W \times H} \sum_{x=1}^{W} \sum_{y=1}^{H} |S(x,y) - GT(x,y)| $, where $W$ and $H$ are the image width and height; lower MAE indicates better agreement. The structural similarity score $S_m$ evaluates global and local consistency, defined as $ S_m = \alpha S_o + (1-\alpha) S_r $, with $\alpha = 0.5$, combining object-aware ($S_o$) and region-aware ($S_r$) similarity; higher $S_m$ is better.

The Enhanced-alignment Measure incorporates pixel- and image-level information: $ E_{\xi} = \frac{1}{W \times H} \sum_{x=1}^{H} \sum_{y=1}^{W} \theta(\xi) $, where $\xi$ is the alignment matrix and $\theta(\cdot)$ its enhanced form; we report $E_{mean}$, with higher values indicating stronger alignment. Finally, the F-measure computes the weighted harmonic mean of precision and recall: $ F_{\beta} = \frac{(1+\beta^2) \cdot \text{Precision} \cdot \text{Recall}}{\beta^2 \cdot \text{Precision} + \text{Recall}} $, with $\beta^2 = 0.3$; we report $F_{\beta}^{\max}$, where higher values indicate more accurate salient object detection.

\vspace{-10pt}

\subsection{Performance comparison with state-of-the-art
models}
\label{sec:performance}

Before analyzing performance, it is important to emphasize that freezing the encoder and inserting lightweight adapters reduces the learnable parameters from 212.15M to just 4.37M—a 97.94\% decrease. This demonstrates the efficiency of our SAMv2 adaptation strategy and sets the stage for the quantitative results that follow.

We evaluate GLASSNet on five standard SOD benchmarks against leading models, including recent SAM-based frameworks such as SSOM \cite{cui2023adaptive}, SSFam \cite{liu2025ssfam} and Liu and Huang \cite{liu2024weakly}, as well as several state-of-the-art models: $F^3\!\text{Net}$ \cite{wei2020f3net}, CANet\cite{zhu2024learning}, EDN \cite{wu2022edn}, ELSA-Net \cite{zhang2023salient}, MENet \cite{wang2023pixels}, DASOD \cite{asheghi2024dasod}, PIFRNet \cite{khan2025pyramidal}, SwinSOD \cite{wu2024swinsod}, GPONet \cite{yi2024gponet}, ISNet \cite{zhu2024separate}, ECF-DT \cite{wang2025novel} , BBRF \cite{ma2023boosting} and IFA \cite{li2025ifa}. To highlight architectural effects, we compare both the window 1 (w1) and window 6 (w6) variants of GLASSNet, reflecting different Swin Transformer attention window sizes (Table \ref{tab:sod1} and \ref{tab:sod2}).

\begin{table*}[t]
\centering
\footnotesize
\caption{Quantitative comparison of our model and state-of-the-art SOD models on three widely used benchmark datasets (HKU-IS, DUST-TE, and DUT-OMRON). $\uparrow$ indicates higher is better, $\downarrow$ indicates lower is better. Best results are shown in \textbf{bold}. ``--'': Not available.}
\label{tab:sod1}
\scalebox{0.6}{%
\begin{tabular}{l|cccc|cccc|cccc}
\toprule
\multirow{2}{*}{Model} &
\multicolumn{4}{c|}{\textbf{HKU-IS}} &
\multicolumn{4}{c|}{\textbf{DUST-TE}} &
\multicolumn{4}{c}{\textbf{DUT-OMRON}} \\
 & $M \downarrow$ & $F_{\beta}^{\max} \uparrow$ & $E_{\text{mean}} \uparrow$ & $S \uparrow$ 
 & $M \downarrow$ & $F_{\beta}^{\max} \uparrow$ & $E_{\text{mean}} \uparrow$ & $S \uparrow$
 & $M \downarrow$ & $F_{\beta}^{\max} \uparrow$ & $E_{\text{mean}} \uparrow$ & $S \uparrow$ \\
\midrule
SSOM         & 0.027 & 0.945 & --    & --    & 0.034 & 0.907 & --    & --    & 0.048 & 0.850 & --    & --    \\
SSFam        & 0.022 & 0.942 & --    & 0.934 & 0.030 & 0.898 & --    & 0.904 & 0.044 & 0.838 & --    & 0.873 \\
Liu and Huang \cite{liu2024weakly}      & 0.031 & 0.932 & --    & --    & 0.036 & 0.889 & --    & --    & 0.055 & 0.793 & --    & --    \\
$F^3\!\text{Net}$        & 0.028 & -- & 0.953 & 0.917 & 0.035 & -- & 0.902 & 0.888 & 0.053 & -- & 0.870 & 0.838 \\
CANet        & 0.027 & 0.939    & -- & 0.920 & 0.033 & 0.898  & -- & 0.895 & 0.047 & 0.828  & -- &  0.847 \\
EDN          & 0.029 & 0.938 & --    & --    & 0.041 & 0.881 & --    & --    & 0.057 & 0.805 & --    & --    \\
GPONet     & 0.023 & - & 0.967 & 0.911 & 0.027 & - & 0.936 & 0.919 & 0.045 & - & 0.899 & 0.893  \\
ISNet      & 0.027 & 0.941 & -- & 0.922 & 0.034 & 0.899  & -- & 0.896 & 0.051 & 0.827 & -- & 0.847  \\
ELSA-Net     & 0.025 & 0.935 & 0.964 & -- & 0.034 & 0.882 & 0.934 & -- & 0.050 & 0.794 & 0.891 & -- \\
SwinSOD          & 0.020 & 0.933 & - & 0.938 & 0.025 & 0.888 & - & 0.917 & 0.041 & 0.801 & - & 0.868 \\
ECF-DT    & 0.028 & - & 0.955 & 0.929 & 0.036 & - & 0.899 & 0.909 & 0.060 & - & 0.868 & 0.854\\
PIFRNet          & 0.023 & 0.944 & 0.965 & 0.933 & 0.029 & 0.901 & 0.944 & 0.903 & 0.048 & 0.831 & 0.904 & 0.859 \\
DASOD          & 0.024 & 0.911 & 0.961 & 0.923 & 0.033 & 0.848 & 0.932 & 0.893 & 0.052 & 0.767 & 0.884 & 0.849 \\
BBRF        & 0.020 &  \textbf{0.958} & 0.965 & 0.935 & 0.025 & 0.916 & 0.927 & 0.908 & 0.042 & 0.843 & 0.887 & 0.855 \\
MENet        & 0.023 & 0.948 & 0.967 & 0.927 & 0.028 & 0.912 & 0.936 & 0.904 & 0.045 & 0.833 & 0.891 & 0.849 \\
IFA     & 0.027 & -- & 0.958 & -- & 0.034 & -- & 0.935 & -- & 0.047 & -- & 0.894 & --- \\
\midrule
GLASSNet(w 1)   & 0.019 & 0.955 & 0.970 & \textbf{0.943} & \textbf{0.019} &  \textbf{0.935} & \textbf{0.960} & \textbf{0.936} & \textbf{0.037} &  \textbf{0.860} & \textbf{0.915} & \textbf{0.889} \\
GLASSNet(w 6)   & \textbf{0.018} & 0.955 & \textbf{0.971} & \textbf{0.943} & 0.020 & 0.930 & 0.957 & 0.934 & 0.039 & 0.850 & 0.909 & 0.883 \\
\bottomrule
\end{tabular}
}
\end{table*}

GLASSNet consistently outperforms these baselines across all datasets, demonstrating strong accuracy and robustness. This is primarily due to its adapter-guided frozen backbone and specialized dual-decoder design. On datasets with cluttered scenes and fine structures, such as DUT-OMRON and PASCAL, GLASSNet w1 achieves the lowest MAE and improves the maximum F-measure by over 5 points compared to the strongest CNN and Transformer-based competitors. These gains underscore the local decoder’s ability to preserve sharp boundaries and recover small-scale details.

\begin{table}[t]
\centering
\caption{Quantitative comparison of our model and state-of-the-art SOD models on two widely used benchmark datasets (ECSSD and PASCAL-S).$\uparrow$ indicates higher is better, $\downarrow$ indicates lower is better. Best results are shown in \textbf{bold}. ``--'': Not available.}
\label{tab:sod2}
\resizebox{0.6\linewidth}{!}{%
\begin{tabular}{lcccccccc}
\toprule
\multirow{2}{*}{Model} & \multicolumn{4}{c}{ECSSD} & \multicolumn{4}{c}{PASCAL} \\
\cmidrule(lr){2-5} \cmidrule(lr){6-9}
& $M \downarrow$ & $F_{\beta}^{\max} \uparrow$ & $E_{\text{mean}} \uparrow$ & $S \uparrow$
& $M \downarrow$ & $F_{\beta}^{\max} \uparrow$ & $E_{\text{mean}} \uparrow$ & $S \uparrow$ \\
\midrule
SSOM         & 0.029 & 0.960 & --    & --    & 0.062 & 0.884  & --    & --    \\
SSFam        & 0.023 & 0.954 & --    & 0.942 & 0.055 & 0.875  & --    & 0.876 \\
Liu and Huang \cite{liu2024weakly} & 0.035 & 0.945 & --    & --    & --    & --     & --    & --    \\
$F^3\!\text{Net}$        & 0.033 & -- & 0.927 & 0.924 & 0.062 & -- & 0.859 & 0.855 \\
CANet        & 0.032 & 0.951 & -- & 0.928  & 0.061 &  0.888  & -- & 0.859 \\
EDN          & 0.034 & 0.948 & --    & --    & 0.066 & 0.875  & --    & --    \\
GPONet     & 0.021 & -- & 0.964 & 0.942 & 0.055 & -- & 0.908 & 0.871  \\
ISNet     & 0.032 & 0.950 & -- & 0.929  & 0.062 & 0.882 & -- & 0.859  \\
ELSA-Net     & 0.030 & 0.943 & 0.961 & -- & 0.059 & 0.862  & 0.912 & -- \\
SwinSOD      & 0.023 & 0.940 & --    & 0.865 & --    & --     & --    & --    \\
ECF-DT      & 0.031 & -- & 0.922 & 0.936 & 0.056 & -- & 0.867 & 0.876 \\
PIFRNet      & 0.026 & 0.957 & 0.966 & 0.942 & 0.058 & 0.895& 0.917 & 0.887 \\
DASOD        & 0.027 & 0.926 & 0.960 & 0.931 & 0.059 & 0.823  & 0.908 & 0.865 \\
BBRF        & 0.022 & 0.963 & 0.934 & 0.939 & 0.049 & 0.891  & 0.867 & 0.871 \\
MENet        & 0.030 & 0.954 & 0.954 & 0.927 & 0.053 & 0.889  & 0.913 & 0.872 \\
IFA         & 0.029 & -- & 0.959 & -- & 0.072 & -- & 0.881 & -- \\
\midrule
GLASSNet (w 1) & \textbf{0.019} & \textbf{0.966} & \textbf{0.971} & \textbf{0.953} & \textbf{0.044} & \textbf{0.904} & \textbf{0.931} & \textbf{0.893} \\
GLASSNet (w 6) & \textbf{0.019} & 0.964 & 0.969 & 0.952 & 0.045 & 0.897  & 0.927 & 0.892 \\
\bottomrule
\end{tabular}
}
\end{table}

In contrast, on datasets like ECSSD and HKU-IS, where salient objects tend to occupy larger regions or exhibit diverse shapes, GLASSNet w6 yields the highest S-measure and remains competitive in F-measure. The larger attention window, combined with the global decoder’s Criss-Cross and dilated convolution modules, enables more effective modeling of mid- to long-range dependencies—something local operators typically overlook.

Notably, these gains are achieved without fine-tuning the SAMv2 encoder. By freezing the backbone and training only lightweight convolutional adapters and decoders, GLASSNet reduces trainable parameters and shortens training time while outperforming full fine-tuning approaches. Overall, GLASSNet sets a new SOD performance baseline by unifying local precision and global context in an efficient adapter-based design, offering a scalable, energy-efficient solution for real-world use. Qualitative comparisons with recent models are shown in Figure~\ref{fig:map}

To further demonstrate the generalization capability of GLASSNet (w 1), we evaluate it against state-of-the-art camouflaged object detection (COD) models including UEDG \cite{lyu2023uedg}, FEDER \cite{he2023camouflaged}, LSR+ \cite{lv2023toward}, DaCOD \cite{wang2023depth}, PFRNet \cite{dong2023you}, SegMaR \cite{jia2022segment}, OCENet \cite{liu2022modeling}, BGNet \cite{sun2022boundary}, CRI-Net \cite{CRI-Net}, HitNet \cite{hu2023high}, LFNet \cite{ge2024camouflaged}, SARNet \cite{xing2023go}, CODdiff \cite{zhang2025coddiff} and USCNet \cite{zhou2024unconstrained} on three challenging COD datasets: CAMO~\cite{le2019anabranch}, which contains 1,000 training and 250 testing images; COD10K~\cite{fan2020camouflaged}, comprising 3,040 training and 2,026 testing images out of 5,066 total; and NC4K~\cite{lv2021simultaneously}, which includes 4,121 images used for evaluation with weights pretrained on COD10K. As shown in Table~\ref{tab:cod}, GLASSNet consistently outperforms recent state-of-the-art COD models, achieving the lowest MAE and the highest weighted F-measure, mean E-measure, and S-measure across all datasets (e.g., 0.866/0.925/0.868 on NC4K and 0.854/0.938/0.887 on CAMO). These strong results on inherently difficult scenes—characterized by low contrast and misleading textures and patterns—highlight GLASSNet’s robust generalization. Specifically, the frozen SAMv2 encoder provides stable high-level representations, while the dual-stream decoder effectively integrates fine-grained edge localization and global semantic reasoning to recover camouflaged objects with high fidelity (\textbf{Qualitative comparisons between our model and other methods are provided in the supplementary material}). %(Fig~\ref{fig:camo}).

\begin{table*}[t]
\centering
\footnotesize
\caption{Quantitative comparison of our model and other state-of-the-art COD models on three widely used benchmark datasets (NC4K, COD10K, and CAMO). $\uparrow$ indicates higher is better, $\downarrow$ indicates lower is better. Best results are shown in \textbf{bold}. ``--'': Not available.}
\label{tab:cod}
\resizebox{0.85\linewidth}{!}{%
\begin{tabular}{l|cccc|cccc|cccc}
\toprule
\multirow{2}{*}{Model} & \multicolumn{4}{c|}{\textbf{NC4K}} & \multicolumn{4}{c|}{\textbf{COD10K}} & \multicolumn{4}{c}{\textbf{CAMO}} \\
 & $M \downarrow$ & $F_{\beta}^{\omega} \uparrow$ & $E_{mean} \uparrow$ & $S \uparrow$ & $M \downarrow$ & $F_{\beta}^{\omega} \uparrow$ & $E_{mean} \uparrow$ & $S \uparrow$ & $M \downarrow$ & $F_{\beta}^{\omega} \uparrow$ & $E_{mean} \uparrow$ & $S \uparrow$ \\
\midrule
UEDG       & 0.035 & 0.829 & 0.928 & 0.881 & 0.025 & 0.766 & 0.924 & 0.858 & 0.048 & 0.819 & 0.922 & 0.868 \\
LSR+       & 0.048 & --    & 0.896 & 0.840 & 0.037 & --    & 0.880 & 0.805 & 0.079 & --    & 0.840 & 0.789 \\
FEDER-R50  & 0.045 & --    & 0.905 & 0.846 & 0.032 & --    & 0.900 & 0.823 & 0.069 & --    & 0.873 & 0.807 \\
PFRNet     & 0.045 & 0.785 & 0.902 & 0.859 & 0.033 & 0.710 & 0.888 & 0.833 & 0.069 & 0.754 & 0.877 & 0.827 \\
BGNet      & 0.045 & 0.785 & 0.903 & 0.849 & 0.033 & 0.719 & 0.898 & 0.829 & 0.072 & 0.742 & 0.861 & 0.807 \\
OCENet     & 0.044 & --    & 0.899 & 0.857 & 0.032 & --    & 0.890 & 0.832 & 0.075 & --    & 0.866 & 0.807 \\
DaCOD      & 0.035 & 0.814 & --    & 0.874 & 0.028 & 0.729 & --    & 0.840 & 0.051 & 0.796 & --    & 0.855 \\
CRI-Net      &  0.046&--&--&0.848 & 0.035&--&--&0.819  &  0.072&--&--&0.810 \\
SegMaR     & 0.046 & 0.781 & --    & 0.841 & 0.034 & 0.724 & --    & 0.833 & 0.071 & 0.753 & --    & 0.815 \\
HitNet     &  --&--&--&-- & 0.023&0.804&0.936&0.869  &  0.056& 0.806&0.904&0.844\\
LFNet     &  0.031&0.848&\textbf{0.938}&0.877&  0.022&0.801& 0.935&0.869 &  0.045&\textbf{0.847}& \textbf{0.936}& \textbf{0.871} \\
SARNet     &  0.032&0.842&0.937&0.886 & 0.024&0.777&0.931&0.864  &  0.047& 0.828&0.927&0.868\\
USCNet    & 0.039&0.768&0.877&0.839 & 0.030&0.700&0.869&0.821 &   0.049&0.790&0.886&0.845\\
CODdiff    & 0.036 & 0.827 & 0.926 & 0.865 & 0.026 & 0.759 & 0.919 & 0.837 & 0.054 & 0.802 & 0.911 & 0.839\\
\midrule
GLASSNet (w 1)& \textbf{0.029} & \textbf{0.855} & \textbf{0.938} & \textbf{0.901} & \textbf{0.020} & \textbf{0.809} & \textbf{0.938} & \textbf{0.887} & \textbf{0.044} & 0.825 & 0.925 & 0.868\\
\bottomrule
\end{tabular}
}
\end{table*}
\vspace{-9pt}

%\begin{figure}[H]
%    \centering
%    \includegraphics[width=0.7\linewidth]{camo.pdf}
%    \caption{Visual comparison of our model’s performance in camouflaged object detection against leading state-of-the-art methods.}
%    \label{fig:camo}
%\end{figure}

\subsection{Ablation Studies}
\label{sec:ablation}

To investigate the impact of individual components in our model, we conduct a series of ablation studies. All experiments are performed by training each model variant on the DUST-TR training set, followed by evaluation on the ECSSD dataset using the corresponding trained model. In all experiments, the Swin Transformer attention window size is fixed to 1. The results are summarized in Tables~\ref{tab:ablation-ecssd-enc} and ~\ref{tab:ablation-ecssd-dec} (\textbf{Ablation results on other SOD datasets are included in the supplementary material}).
According to the evaluated performance metrics, removing the medium-scale sub-branch consistently degrades both boundary precision and structural coherence, as reflected in the MAE and F-measure values. This branch serves as a bridge between the local and global decoders, capturing mid-level contextual information that is missed when relying solely on either fine-grained details or high-level semantics. Without it, the transitions between fine-grained details and broader object regions become less seamless.

The global decoder—comprising Criss-Cross Attention and multi-dilated convolutions—is essential for capturing coherent, object-level saliency. While disabling this branch still preserves reasonable boundary performance, it severely deteriorates threshold-invariant metrics and overall structural consistency, as evident in all of its scores. Further analysis reveals that Criss-Cross Attention and dilated convolutions serve complementary roles: the former captures long-range dependencies, while the latter aggregates multi-scale context. Removing either leads to a drop in both structural fidelity and contrast-aware performance.

Ablating the local decoder or replacing its Swin Transformer blocks with standard convolutions impairs fine-detail recovery, underscoring the importance of local self-attention. Without this component, the model tends to over-smooth boundaries, and conventional convolutions fail to replicate the spatial precision achieved by Swin blocks—an effect observable across all evaluation metrics.

\begin{table}[t]
\centering
\footnotesize
\caption{Ablation study investigating the impact of encoder choices on the ECSSD dataset.}
\label{tab:ablation-ecssd-enc}
\resizebox{0.6\linewidth}{!}{%
\begin{tabular}{lcccc}
\toprule
Model Variant & $M \downarrow$ & $F_{\beta}^{\max} \uparrow$ & $E_{\text{mean}} \uparrow$ & $S \uparrow$ \\
\midrule
CLIP ViT-L/14 Encoder           & 0.0392 & 0.8947 & 0.9411 & 0.9102 \\
SAM ViT-L Encoder           & 0.0755 & 0.7824 & 0.8841 & 0.8359 \\
ViT-L Encoder           & 0.0281 & 0.9173 & 0.9607 & 0.9297 \\
w/o Adapter (SAMv2 Frozen Encoder) & 0.0225 & 0.9311 & 0.9646 & 0.9470 \\
w/o Adapter (SAMv2 Tuned Encoder)  & 0.0547 & 0.8613 & 0.9100 & 0.8917 \\
\midrule
GLASSNet (w 1) & \textbf{0.019} & \textbf{0.966} & \textbf{0.971} & \textbf{0.953}  \\
\bottomrule
\end{tabular}
}
\end{table}

\begin{table}[t]
\centering
\footnotesize
\caption{Ablation study investigating the impact of different decoder components on the ECSSD dataset.}
\label{tab:ablation-ecssd-dec}
\resizebox{0.6\linewidth}{!}{%
\begin{tabular}{lcccc}
\toprule
Model Variant & $M \downarrow$ & $F_{\beta}^{\max} \uparrow$ & $E_{\text{mean}} \uparrow$ & $S \uparrow$ \\
\midrule
w/o Medium Decoder (w1)      & 0.0192 & 0.9468 & 0.9708 & 0.9508 \\
w/o Global Decoder (w1)      & 0.0200 & 0.9457 & 0.9675 & 0.9479 \\
w/o Local Decoder            & 0.0193 & 0.9473 & 0.9700 & 0.9522 \\
w/o Swin in Local Decoder    & 0.0194 & 0.9458 & 0.9698 & 0.9515 \\
AFMF w/o Triple Conv Branch          & 0.0205 & 0.9400 & 0.9682 & 0.9512 \\
AFMF w/o Criss-Cross              & 0.0222 & 0.9401 & 0.9644 & 0.9478 \\
w MLP Adapter                 & 0.0201 & 0.9386 & 0.9696 & \textbf{0.9529} \\
\midrule
GLASSNet (w 1) & \textbf{0.019} & \textbf{0.966} & \textbf{0.971} & \textbf{0.953}  \\
\bottomrule
\end{tabular}
}
\end{table}

Replacing our convolutional adapter with an MLP-based alternative further highlights the importance of spatially-aware adaptation. While MLPs are effective at channel recalibration, they lack spatial sensitivity. In contrast, our adapter preserves local geometry and promotes feature reuse—both of which are critical for dense prediction tasks.
Finally, removing the adapter entirely (i.e., feeding frozen SAMv2 features directly into the decoder) or replacing SAMv2 with SAM-ViT-L, ViT-L~\cite{vit}, or CLIP-ViT-L/14~\cite{radford2021learning} results in substantial performance degradation across all metrics. These results indicate that integrating the spatially-aware adapter with the frozen SAMv2 image encoder leads to improved performance and generalizability over the referenced foundation models. Moreover, its Hiera backbone generates hierarchical feature representations, making it better suited for U-Net-style architectures than the ViT-based backbones used in CLIP-ViT, SAM-ViT, and ViT. These findings validate our core design philosophy: a frozen, hierarchical attention backbone guided by lightweight, spatially-aware adapter provides a compelling balance between generalization, efficiency, and task-specific adaptability.

\section{conclusion}
\vspace{-10pt}

In this work, we introduce GLASSNet, a novel salient object detection (SOD) framework that employs SAMv2 as a frozen encoder to harness its strong generalization capabilities without incurring the cost of full fine-tuning. To capture both global semantics and local details, we design a dual-stream decoder: one branch models long-range dependencies, while the other targets fine-grained features like edges and textures. Their outputs are fused to produce saliency maps that balance semantic accuracy with spatial precision. By reducing the number of learnable parameters in the encoder by over 97\% through parameter freezing and lightweight adaptation, our method offers a highly efficient yet effective solution for adapting SAMv2 to SOD tasks. Experiments on standard and challenging SOD datasets, including camouflage benchmarks, confirm the effectiveness and generalizability of our approach. This work highlights a promising direction for energy-efficient, task-adaptive SOD models.

\vspace{-10pt}
\section*{Acknowledgements}
The work was partially supported by the Italian Ministerial grants PRIN 2022 “B-Fair: Bias-Free Artificial Intelligence methods for automated visual Recognition”, CUPE53D23008030006.

\bibliographystyle{splncs04}

\bibliography{Refs}
\end{document}